\documentclass[10pt,twocolumn,letterpaper]{article}
\pdfoutput=1
\usepackage[pagenumbers]{cvpr} 

\usepackage[dvipsnames,table]{xcolor}

\definecolor{cvprblue}{rgb}{0.21,0.49,0.74}
\usepackage[pagebackref,breaklinks,colorlinks,citecolor=cvprblue]{hyperref}

\usepackage{tabularx}  

\usepackage{subfiles}
\usepackage{amsmath}
\usepackage{mathtools}
\usepackage[normalem]{ulem}
\usepackage{multirow}
\usepackage{diagbox}
\usepackage{wrapfig}
\usepackage{verbatim}
\usepackage{amssymb}
\usepackage{pifont}
\usepackage{tikz}
\usepackage{comment}
\usepackage{soul}
\sethlcolor{red}

\def\paperID{7144} 
\def\confName{CVPR}
\def\confYear{2024}

\title{Language-Driven Visual Consensus for Zero-Shot Semantic Segmentation}

\author{%
    Zicheng Zhang$^{1}$ \quad
    Tong Zhang$^{4}$ \quad
    Yi Zhu$^{2}$ \quad
    Jianzhuang Liu$^{2}$ \quad \\
    Xiaodan Liang$^{3}$ \quad
    QiXiang Ye$^{5}$ \quad
    Wei Ke$^{1}$ \quad \\[5pt]
    $^1$School of Software Engineering, Xi'an Jiaotong University \\
    $^2$Huawei Noah's Ark Lab \quad $^3$Sun Yat-sen University \quad
    $^4$  EPFL \\
    $^5$ University of Chinese Academy of Sciences
}

\begin{document}
\maketitle
\begin{abstract}

The pre-trained vision-language model, exemplified by CLIP~\cite{radford2021learning}, advances zero-shot semantic segmentation by aligning visual features with class embeddings through a transformer decoder to generate semantic masks. Despite its effectiveness, prevailing methods within this paradigm encounter challenges, including overfitting on seen classes and small fragmentation in masks. 
To mitigate these issues, we propose a Language-Driven Visual Consensus (LDVC) approach, fostering improved alignment of semantic and visual information.
Specifically, we leverage class embeddings as anchors due to their discrete and abstract nature, steering vision features toward class embeddings. Moreover, to circumvent noisy alignments from the vision part due to its redundant nature, we introduce route attention into self-attention for finding visual consensus, thereby enhancing semantic consistency within the same object. Equipped with a vision-language prompting strategy, our approach significantly boosts the generalization capacity of segmentation models for unseen classes. Experimental results underscore the effectiveness of our approach, showcasing mIoU gains of 4.5\% on the PASCAL VOC 2012 and 3.6\% on the COCO-Stuff 164k for unseen classes compared with the state-of-the-art methods.

\end{abstract}  
\section{Introduction}
\begin{figure}[t]
\centering
        \includegraphics[width=0.99\linewidth]{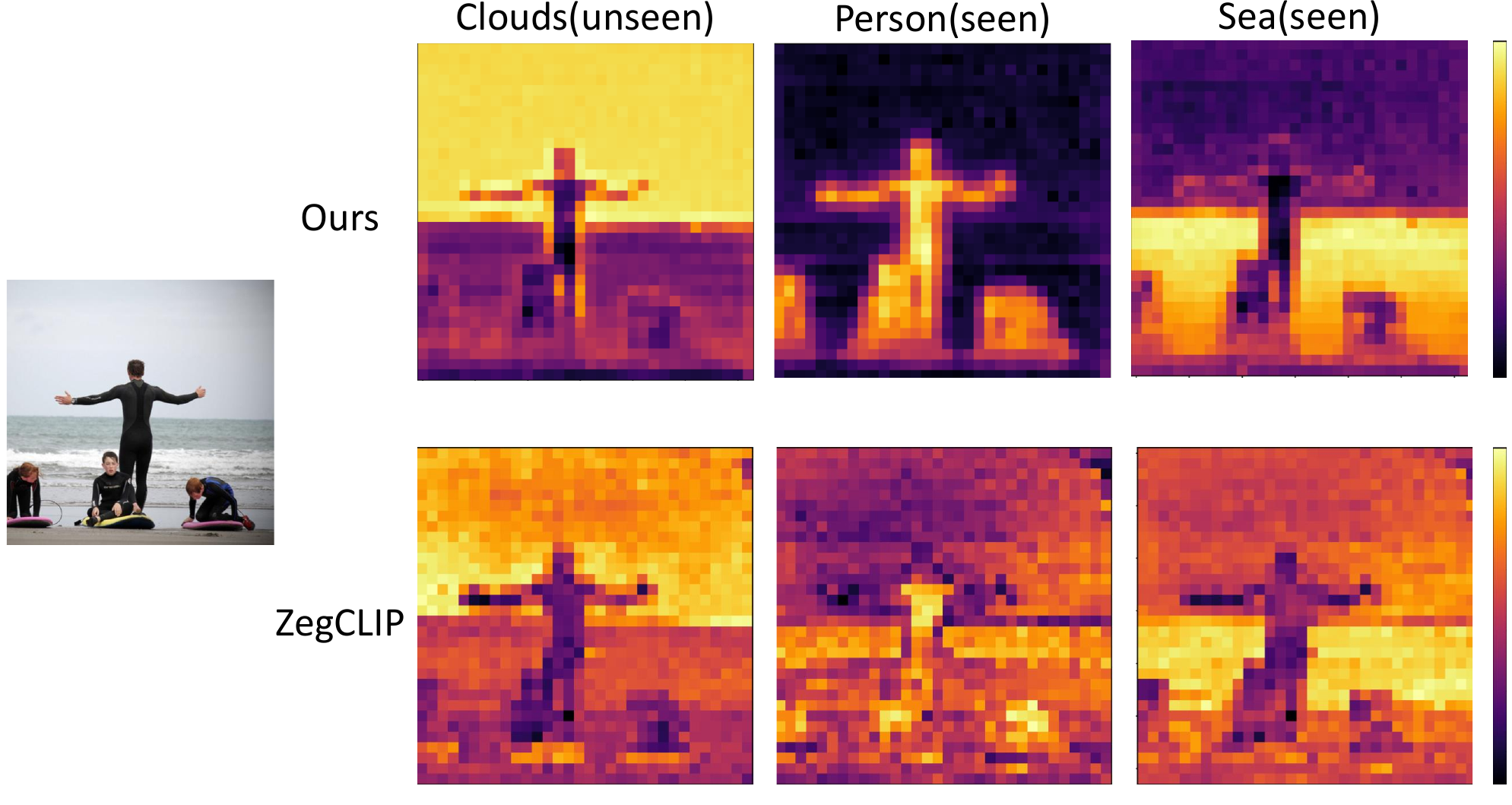}

\caption{We visualize the cross attention maps between image features and class embeddings from last transformer decoder block in both our approach and the SOTA method, \textit{i.e.}, ZegCLIP~\cite{zhou2022zegclip}. It shows that our approach establishes a clearer, more explicit alignment between dense image features and class embeddings.}
\label{fig:alignment comparing}
\vspace{-1.0em}
\end{figure}

The past decade has witnessed the tremendous success of deep learning methods~\cite{kirillov2023segment,zheng2021rethinking,ranftl2021vision,strudel2021segmenter,cheng2021per,zhang2022segvit,cheng2022masked} in the field of semantic segmentation. However, limited to labor-intensive pixel-wise annotations, it is difficult to extend these methods to scenarios with a greater number of object categories. This hinders their application in open-world scenarios, such as autonomous driving~\cite{hu2023planning,hu2023gaia} and embodied intelligence~\cite{gupta2021embodied,roy2021machine}. To overcome this problem, \textbf{Z}ero-\textbf{S}hot \textbf{S}emantic \textbf{S}egmentation (ZS3)~\cite{bucher2019zero} is proposed to extend the capability of aligning pixels with pre-defined classes to novel unseen classes.

Recently, the emergence of vision-language pre-trained models~\cite{radford2021learning,jia2021scaling,li2023scaling,xu2023demystifying,cherti2023reproducible} boost the development of ZS3 due to their remarkable image-level zero-shot capability. Consequently, the primary challenge in ZS3 has transitioned to transferring this zero-shot capability from the image level to the pixel level.  Pioneering works~\cite{xu2022simple,ding2022decoupling,qin2023freeseg} initially introduced a two-stage method, employing a mask proposal network for mask generation and CLIP as an open-vocabulary classifier for mask labeling. Subsequent endeavors~\cite{kim2023zegot,zhou2022zegclip,zhou2022maskclip,rao2022denseclip,li2022languagedriven} propose a one-stage method to enhance inference speed while maintaining or improving segmentation performance. It equips CLIP model with an extra  deocoder for further tuning alignment in pixel level. Among those works, ZegCLIP~\cite{zhou2022zegclip} underscores the importance of fine-tuning strategies. Compared to full fine-tuning and parameter-free tuning, parameter-efficient tuning can achieve a trade-off between maintaining zero-shot ability from CLIP and better segmentation performance. 

\begin{figure}[!t]
\centering
        \includegraphics[width=0.99\linewidth]{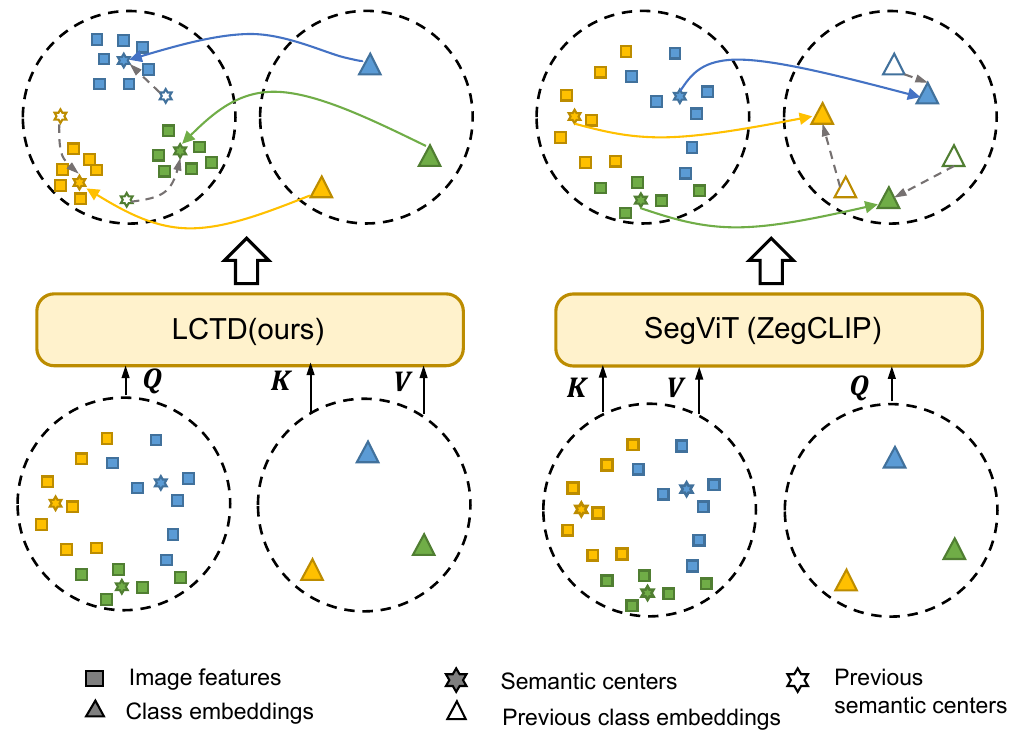}

\caption{\textbf{Illustration of difference between Our Decoder and ZegCLIP's Decoder~\cite{zhou2022zegclip}.} We depict the change of the visual and semantic spaces before and after the decoder. Our approach initially contracts the region of the visual space, maximizing its similarity to the class embeddings. In contrast, ZegCLIP~\cite{zhou2022zegclip} pushes class embeddings towards noisy and redundant visual cues, resulting in significant semantic drift.}
\vspace{-0.5em}
\label{fig:introduce}
\vspace{-1.0em}
\end{figure}

Despite these advancements, the one-stage method still encounters with overfitting on seen classes~\cite{li2023tagclip}, attributed to the lack of appropriate modifications in the transformer decoder~\cite{carion2020end}, such as SegViT~\cite{zhang2022segvit}. This line of works commonly selects class embeddings from CLIP's text encoder as query and image features as key and value in cross attention. While effective in fully supervised semantic segmentation, this paradigm falters in zero-shot semantic segmentation, especially when introducing novel classes during inference. This stems from the fact that the visual space from CLIP is not structured enough for segmentation~\cite{zhou2022maskclip}. Moreover, parameter-free/efficient fine-tuning introduces constraints on updates. Considering the redundant nature of visual cues, pushing class embeddings toward visual features can dramatically shift the existing semantic space, compromising zero-shot ability, as depicted in Fig.~\ref{fig:introduce}. In contrast, language is a more abstract and highly structured concept, especially in the context of fine-tuning on the CLIP model. This inherent richness in linguistic representation enables the model to capture diverse visual information encapsulated by language.  Consequently, making subtle adjustments to the semantic space facilitates the model's adaptability to new datasets while preserving the robust generalization ability inherited from CLIP.\label{introduce:explain}


Building upon this intuition, we present our Language-Driven Visual Consensus (LDVC) approach, designed to perform comprehensive fine-tuning on both visual and language prompts. In this approach, we utilize language representations as anchors to guide the refinement of visual cues. Specifically, we implement a straightforward substitution where image features act as query, and class embeddings serve as both key and value in the cross-attention mechanism of the transformer decoder. This substitution serves a dual purpose: it ensures the consistency of class embeddings and contributes to the creation of a more compact visual space. Additionally, we introduce a Local Consensus Transformer Decoder (LCTD) to function as the pixel alignment decoder, updating image features under the guidance of class embeddings. To address noise introduced by the redundancy in visual cues, we apply a route attention mechanism, effectively alleviating small fragmentation in segmentation masks before their alignment with class embeddings. The compactness of the visual space not only enhances the alignment process but also facilitates effective language tuning, as depicted in Fig.~\ref{fig:introduce}. This strategic adjustment serves to mitigate the risk of semantic drift and promotes the development of a more stable and structured semantic space during fine-tuning. Consequently, this fortifies the zero-shot capability of the model, making it more resilient and adaptable across diverse datasets.

In a nutshell, our contributions can be summarized as:
\begin{itemize}
    \item A new local consensus transformer decoder is proposed to alleviate overfitting on unseen classes and reduce the small fragmentation in masks.
    \item A vision-language prompt tuning strategy is proposed to generalize the pre-trained CLIP to zero-shot semantic segmentation, which further improves unseen classes segmentation ability.
\end{itemize}


To validate the effectiveness of our approach, we conduct experiments on the publicly available PASCAL VOC 2012 and COCO-Stuff 164k datasets. Results demonstrate that our approach outperforms the state-of-the-art method by 0.6\% for seen classes and 4.5\% for unseen classes on the PASCAL VOC 2012 dataset, as well as 3.0\% and 3.6\% on the COCO-Stuff 164k dataset, respectively.

\section{Related Works}
\noindent\textbf{Zero-Shot Semantic Segmentation.} Despite the significant success of deep learning methods~\cite{kirillov2023segment,zheng2021rethinking,ranftl2021vision,strudel2021segmenter,cheng2021per,zhang2022segvit,cheng2022masked} in image segmentation, the difficulty in obtaining pixel-wise annotations has constrained the generalization of these approaches. To overcome this problem, researchers propose zero-shot semantic segmentation~\cite{bucher2019zero}. Nowadays, depending on the granularity of zero-shot, this task develop two streams. One stream~\cite{xu2022groupvit,cai2023mixreorg,ren2023viewco,luo2023segclip,chen2023exploring,ghiasi2022scaling,minderer2205simple} focuses on how to learn region-text alignment during pretraining stage, which eases the demand for pixel-level annotations. The other stream focuses on transferring the alignment capability of image-text pretrained models~\cite{radford2021learning,jia2021scaling,li2023scaling,xu2023demystifying,cherti2023reproducible} to the pixel-text level during fine-tuning stage. Recent works in this stream derive two methods: two-stage method and one-stage method. Two-stage method~\cite{xu2022simple,ding2022decoupling,qin2023freeseg} decouples mask generation and mask classification. It first utilizes a mask proposal network to generate class-agnostic masks and then leverages a pretrained image-text model, like CLIP~\cite{radford2021learning} to classify the masked regions in image. Although it well maintains the zero-shot capabilities of CLIP, the inference cost is heavy due to the separate classification of the huge amount of masked regions per image. The one-stage method~\cite{kim2023zegot,zhou2022zegclip,zhou2022maskclip,rao2022denseclip,li2022languagedriven,Han_2023_ICCV,ding2023openvocabulary,han2023open,jiao2023learning}, however, directly finetunes the pretrained text-image model or distills its knowledge to a side network on segmentation datasets. One-stage method has lower inference costs but sacrifices some degree of alignment capabilities of the pretrained model. Our new local consensus transformer decoder effectively addresses the issue of overfitting on seen classes in one-stage methods.

\noindent\textbf{Prompt tuning.} 
Prompt tuning was first introduced in natural language processing~\cite{brown2020language,li2021prefix,lester2021power}. This technique freezes the parameters of pretrained language model and  utilizes additional tokens to facilitate fine-tuning of language models. This new fine-tuning strategy, while adapting to downstream tasks, better avoids ``catastrophic forgetting"~\cite{goodfellow2013empirical}. Due to its effectiveness, recent works try to introduce this technique into vision language model learning. The approaches in this field can be divided into three aspects: 1) vision-only prompt tuning, 2) language-only prompt tuning, and 3) vision-language prompt tuning.

Vision-only prompt tuning mainly focuses on  fine-tuning of pretrained ViT~\cite{dosovitskiy2020image}. 
VPT~\cite{jia2022visual} and EXPRES~\cite{das2023learning} insert additional learnable tokens between image patch tokens and [CLS] token in each block during fine-tuning. Language-only prompt tuning~\cite{zhu2023prompt,zhou2022learning,zhou2022conditional,rao2022denseclip,lu2022prompt} focuses on optimizing downstream text prompts close to the format of text used for CLIP~\cite{radford2021learning} pretraining. CoOp~\cite{zhou2022learning} first introduced learnable prompt toekns in text embeddings. Further, CoCoOp~\cite{zhou2022conditional} and DenseCLIP~\cite{rao2022denseclip} use image features to condition learnable prompt tokens for better adaption. Vision-language prompt tuning~\cite{khattak2022maple,lee2023read} inserts learnable prompt tokens in both the visual encoder and the text encoder for CLIP's fine-tuning. We introduces vision-language prompt tuning into zero-shot semantic segmentation task and demonstrates it expands the zero-shot segmentation capability of CLIP.
\section{Methodology}
\begin{figure*}[t]
\centering        \includegraphics[width=0.98\textwidth]{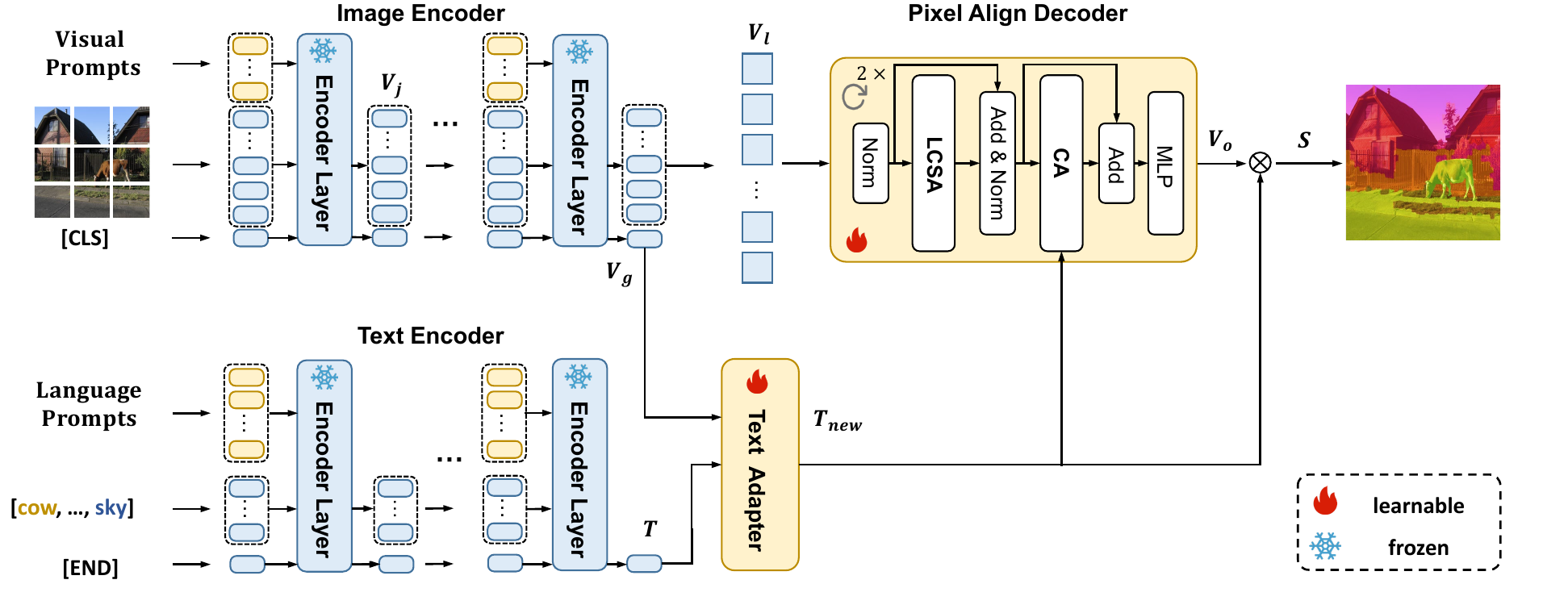}
\caption{\textbf{The architecture of the proposed Language Driven Visual Consensus (LDVC) approach}. We freeze the parameters of CLIP and insert deep learnable visual-language prompts into encoders. Then a text adapter fuses class embeddings $T$ and global image feature $V_g$. At last, the local-consensus transformer decoder progressively updates the dense visual features $V_l$ under the guidance of updated class embeddings $T_{new}$ to get the final segmentation logits $S$.}
\label{fig:arch}
\end{figure*}

In this section, we first provide a brief introduction to the ZS3 setting and review a meta-architecture for the text-image segmentation model. Then the vision-language prompt tuning for CLIP and the new local-consensus transformer decoder are detailed, which are the keys to maintaining and transferring the alignment of CLIP in ZS3.

\subsection{Preliminary}

We follow the generalized zero-shot semantic segmentation(GZS3) protocol~\cite{xian2019semantic}, in which it divides the class set of segmentation datasets into two disjoint subsets: the seen class set $C^s$ and the unseen class set $C^u$. The model training is conducted only on $C^s$ while evaluation on both $C^s$ and $C^u$. Specifically, during the training phase, we only have pixel-wise annotations about $C^s$ on each image. During the inference phase, we need to classify each pixel of the image as one of the classes in $C=C^u \cup C^s$. There are two typical settings for ZS3.
Inductive ZS3 setting
aims to solve the case that unseen class names are not known during training, while transductive ZS3 setting is specified with the unseen class names. Our method mainly focuses on inductive setting, and we also evaluate the performance of transductive setting.

\subsection{Overview Pipeline}

Previous works~\cite{zhao2023unleashing,rao2022denseclip,zhou2022zegclip,zhou2022maskclip} that apply text-image models as backbone to segmentation tasks can be reduced to a simple meta-architecture. This architecture simply casts ZS3 as a pixel-level zero-shot classification problem. This architecture usually consists of three components. CLIP's image encoder and text encoder extracts dense image features and class embeddings respectively. A pixel alignment decoder aligns class embeddings with dense image features. Finally, a simple dot product between updated class embeddings and updated dense image features produces the segmentation mask logits. For this architecture in ZS3 task, the key lies in maintaining the alignment capability of CLIP and transferring this alignment ability to the pixel level. These two aspects correspond to the fine-tuning strategy for CLIP’s encoders and the approach of alignment in pixel- alignment decoder, respectively. 

\subsection{Vision-language Prompt Tuning}
\label{par:VLPT}
Based on the meta architecture mentioned above, we first address the fine-tuning approach for CLIP's encoders. Currently, the approaches for fine-tuning CLIP's encoders can be divide into three aspects. Full fine-tuning methods update all parameters of CLIP. Parameter-free tuning methods freeze all parameters of CLIP and Parameter-efficient tuning methods~\cite{hu2021lora,gao2023clip,zhou2022learning} update small part of parameters in CLIP. As mentioned in ZegCLIP~\cite{zhou2022zegclip}, full fine-tuning methods gain high IoU on seen classes but fail on unseen classes and parameter-free tuning methods fail on seen classes. That means the former undermines CLIP's alignment capability, while the latter cannot enhance CLIP's segmentation capability. Thus, parameter-efficient tuning methods become a trade-off solution in this dilemma. 

Beyond ZegCLIP, we apply deep prompt tuning~\cite{jia2022visual} on both image and text encoders which shows better performance in Tab.~\ref{tab:prompt methods}. As we choose CLIP ViT-B as our default encoders, both of them are transformer-like~\cite{vaswani2017attention} networks which process sequence input. Formally, we divide the sequence embeddings $e_i$ in each layer into three parts:  prompt embeddings $e^{(i)}_{\textrm{p}}$, content embeddings $e^{(i)}_{\textrm{c}}$, global embeddings $e^{(i)}_{\textrm{g}}$,
\begin{align}
    e^{(i)} = [e^{(i)}_{\textrm{p}};e^{(i)}_{\textrm{c}};e^{(i)}_{\textrm{g}}],
\end{align}
where ``[;]'' represents concatenation operation of tensor. $e_p^{(i)}$ is deep learnable prompt embeddings in each layer.
For vision encoder, $e^{(i)}_{\textrm{c}}$ corresponds to image patch embeddings and $e^{(i)}_{\textrm{g}}$ corresponds to [CLS] token embeddings which represent the whole image. For text encoder, $e^{(i)}_{\textrm{c}}$ corresponds to class name embeddings and  $e^{(i)}_{\textrm{g}}$ corresponds to [EOS] token embeddings which represent the whole sentence. For transformer encoder with $l$ layers, the forward of sequence with learnable prompts can be formulated as:
\begin{align}
[\_;e^{(i)}_{\textrm{c}};e^{(i)}_{\textrm{g}}] &= \mathbf{L}_{i} ([e^{(i-1)}_{\textrm{p}};e^{(i-1)}_{\textrm{c}};e^{(i-1)}_{\textrm{g}}]), 
\end{align}
where $\mathbf{L}_{i}$ denotes $i$-th layer of transformer. During the training stage, we freeze all parameters in $\mathbf{L}_{i}$ and only optimize $e_p^{(i)}$.

Through our experiments, we also find that the initialization of $e^{(i)}_p$ has a significant impact on the text encoder. Usually, $e^{(i)}_p$ is initialized from a certain distribution, like uniform or normal distribution. However, we find that the embeddings of hand-crafted prompts would be a better initialization for $e^{(i)}_p$. Specifically, for the input text like \textit{``a photo of a \{class name\}''}, we call the template \textit{``a photo of a ''} as hand-crafted prompts. We first sent these hand-crafted prompts to the pretrained text encoder and extract the corresponding content embeddings $\mathcal{P}^{(i)}$ in each layer. Then we use $\mathcal{P}^{(i)}$ to initialize $e^{(i)}_p$ at the beginning of training. For the prompt embeddings in image encoder, we follow the random initialization as the same as VPT~\cite{jia2022visual}.

\subsection{Local Consensus Transformer Decoder}
\label{par:LCTD}
After determining the fine-tuning approach for CLIP, we review the design of the pixel alignment decoder. In ZegCLIP, it simply adopts SegViT~\cite{zhang2022segvit} as its pixel alignment decoder, which is a transformer decoder with attention-to-mask module. It chooses visual features as key and value and class embeddings as query in cross attention layer. It progressively updates the class embeddings and keeps the image features unchanged during decoding stage. As illustrated in introduction, this is suboptimal under parameter-efficient tuning, leading to a less structured visual space. Thus we propose a new local consensus transformer decoder to mitigate this issue.

Specifically, Given an image $I\in \mathbb{R}^{H\times W \times 3}$ and class sets $C$, the CLIP encoders with visual-language prompts output dense image feature $V_l\in \mathbb{R}^{h\times w\times d}$,
multi-intermediate image features $\{V_j\}_{j=1,2,\dots,N}$, 
global image feature $V_g\in \mathbb{R}^{d}$ (corresponding to [CLS] token), and class embeddins $T \in \mathbb{R}^{c\times d}$, where $h$ and $w$ are the height and width of dense image features respectively, $d$ is the dimension of the CLIP's joint embedding space, $c$ is the number of semantic categories, and $N$ is the number of intermediate image features which is equal to the number of blocks in local consensus transformer decoder. 

Following the design of previous methods~\cite{kim2023zegot,zhou2022zegclip}, we utilize a text adapter to fuse $V_g$ and $T$ to update class embeddings based on visual cues. The text adapter is implemented as:
\begin{align}
    T_{\mathrm{new}} = \mathbf{Proj}([T \odot V_g; T ]),
\end{align}
where $\mathbf{Proj}$ is a linear projection, $\odot$ represents broadcast element-wise multiplication. The fused class embeddings $T_{\mathrm{new}}\in \mathbb{R}^{c\times d}$ have the same size as $T$. 

The local consensus transformer decoder chooses image features as query and fused class embeddings $T_{\mathrm{new}}$ as key and value in cross-attention. Each block in our decoder consists of local consensus self-attention, cross-attention and multi-layer perception. Formally, the block can be expressed as:
\begin{align}
&V  = V_{in} + \textbf{Conv}(V_j) \\
&V  = V + \mathbf{LCSA}(\mathbf{Norm}(V)) \\
&V  = V + \mathbf{CA}(\mathbf{Norm}(V),\mathbf{Norm}(T_{\mathrm{new}})) \\
& V_{out}  = V +\mathbf{MLP}(\mathbf{Norm}(V)),
\end{align}
where $V_{in}$ is the previous block's output feature and $V_{j}$ is the the intermediate image feature
of CLIP encoder, \textbf{Conv} is a convolution layer with $1\times1$ kernel , \textbf{LCSA} is local consensus self-attention, \textbf{Norm} is layer normalization~\cite{ba2016layer}, \textbf{CA} represents cross-attention and \textbf{MLP} is a simple multi-layer perception. 

As semantic segmentation can be viewed as a clustering problem, semantic correspondence in objects is crucial in segmentation decoding stage, which can alleviate the phenomenon of small fragmentation in segmentation mask. To better enhance the semantic correspondence in our decoder, our local consensus self-attention utilizes route attention~\cite{zhu2023biformer,roy2021efficient,cheng2023hybrid} to dynamically select the regions that most semantic-relevant to attend in self-attention layers of the transformer decoder. Specifically, For dense image features $V$, we first employ an average pooling layer with a window size of $n\times n$ to extract window features $V_w\in \mathbb{R}^{h/n\times w/n \times d}$. Then we calculate the similarity between each window feature and choose $m$ windows that are most semantic-relevant for each window feature. Based on the routing, we constrain self-attention for $V$ in the selected $m$ windows. The operations are as follows:
\begin{align}
    V_w & = \mathbf{AvgPool}(V_{in}) \\
    Q,K,V & = \mathbf{Chunk}(\mathbf{Proj}(V_{in})) \\
    \mathrm{ids} & = \mathbf{Topk}(V_w V_w^T,m) \\
    K_{s}, V_{s} &= \mathbf{Gather}(K,\mathrm{ids}), \mathbf{Gather}(V,\mathrm{ids}) \\
    O &= \mathbf{Softmax}(QK_{s}^T/\sqrt{d}) V_{s},
\end{align}
where $\mathbf{AvgPool}$ denotes average pooling operation, $\mathbf{Proj}$ denotes linear projection which projects $V_{in}$ from length $d$ to $3d$, $\mathbf{Chunk}$ means splitting the tensor along the last dimension, $\mathbf{Topk}$ returns the indices of $m$-largest elements of given tensor, $\mathbf{Gather}$ denotes the operation that gathers tensors according to indices, and $\mathbf{Softmax}$ denotes a softmax layer. 

Finally, after forwarding several stacked decoder blocks, we get the final image feature $V_o\in \mathbb{R}^{h\times w \times d}$ and we multiply $V_o$ with $T_{\mathrm{new}}$ to produce segmentation logits $S\in \mathbb{R}^{c\times h \times w}$, as:
\begin{align}
    S =  T_{\mathrm{new}} V_o^T.
\end{align}

\begin{table*}[t]
\centering
\resizebox{0.99\linewidth}{!}{%
\setlength{\tabcolsep}{0.7em}
    {\renewcommand{\arraystretch}{1.1}
\begin{tabular}{cc|cl|cccc|cccc}
\toprule
\multicolumn{2}{l|}{}  &\multicolumn{1}{l|}{\multirow{2}{*}{Pipeline}}      & \multicolumn{4}{c|}{\textbf{PASCAL VOC 2012}} &\multicolumn{4}{c}{\textbf{COCO-Stuff 164k}}
\\ \cline{4-11} 

\multicolumn{2}{l|}{}  & \multicolumn{1}{l|}{} & \multicolumn{1}{c|}{pAcc}            & \multicolumn{1}{c|}{mIoU(S)}          & \multicolumn{1}{c|}{\textbf{mIoU(U)}} &\multicolumn{1}{c|}{hIoU}& \multicolumn{1}{c|}{pAcc}            & \multicolumn{1}{c|}{mIoU(S)}          & \multicolumn{1}{c|}{\textbf{mIoU(U)}} &\multicolumn{1}{c}{hIoU}\\ \hline

\multicolumn{2}{l|}{SPNet~\cite{xian2019semantic}}  & \multicolumn{1}{l|}{One-Stage}     & \multicolumn{1}{c|}{ - } & \multicolumn{1}{c|}{ 78.0 } & \multicolumn{1}{c|}{ 15.6 } & \multicolumn{1}{c|}{ 26.1 } & \multicolumn{1}{c|}{ - } & \multicolumn{1}{c|}{ 35.2 } & \multicolumn{1}{c|}{ 8.7 } & \multicolumn{1}{c}{ 14.0 }         \\

\multicolumn{2}{l|}{ZS3~\cite{bucher2019zero}}  & \multicolumn{1}{l|}{One-Stage}     & \multicolumn{1}{c|}{ - } & \multicolumn{1}{c|}{ 77.3 } & \multicolumn{1}{c|}{ 17.7 } & \multicolumn{1}{c|}{ 28.7 } & \multicolumn{1}{c|}{ - } & \multicolumn{1}{c|}{ 34.7 } & \multicolumn{1}{c|}{ 9.5 } & \multicolumn{1}{c}{ 15.0 }         \\ 

\multicolumn{2}{l|}{CaGNet~\cite{gu2020context}}  & \multicolumn{1}{l|}{One-Stage}     & \multicolumn{1}{c|}{ 80.7 } & \multicolumn{1}{c|}{ 78.4 } & \multicolumn{1}{c|}{ 26.6 } & \multicolumn{1}{c|}{ 39.7 } & \multicolumn{1}{c|}{ 56.6 } & \multicolumn{1}{c|}{ 33.5 } & \multicolumn{1}{c|}{ 12.2 } & \multicolumn{1}{c}{ 18.2 }         \\ 

\multicolumn{2}{l|}{SIGN~\cite{cheng2021sign}} & \multicolumn{1}{l|}{One-Stage}     & \multicolumn{1}{c|}{ - } & \multicolumn{1}{c|}{ 75.4 } & \multicolumn{1}{c|}{ 28.9 } & \multicolumn{1}{c|}{ 41.7 } & \multicolumn{1}{c|}{ - } & \multicolumn{1}{c|}{ 32.3 } & \multicolumn{1}{c|}{ 15.5 } & \multicolumn{1}{c}{ 20.9 }         \\ 

\multicolumn{2}{l|}{STRICT~\cite{pastore2021closer}}  & \multicolumn{1}{l|}{One-Stage}     & \multicolumn{1}{c|}{ - } & \multicolumn{1}{c|}{ 82.7 } & \multicolumn{1}{c|}{ 35.6 } & \multicolumn{1}{c|}{ 49.8 } & \multicolumn{1}{c|}{ - } & \multicolumn{1}{c|}{ 35.3 } & \multicolumn{1}{c|}{ 30.3 } & \multicolumn{1}{c}{ 32.6 }         \\

\multicolumn{2}{l|}{ZegFormer~\cite{ding2022decoupling}}  & \multicolumn{1}{l|}{Two-Stage}     & \multicolumn{1}{c|}{ - } & \multicolumn{1}{c|}{ 86.4 } & \multicolumn{1}{c|}{ 63.6 } & \multicolumn{1}{c|}{ 73.3 } & \multicolumn{1}{c|}{ - } & \multicolumn{1}{c|}{ 36.6 } & \multicolumn{1}{c|}{ 33.2 } & \multicolumn{1}{c}{ 34.8 }         \\

\multicolumn{2}{l|}{Zsseg~\cite{xu2022simple}}  & \multicolumn{1}{l|}{Two-Stage}     & \multicolumn{1}{c|}{ 90.0 } & \multicolumn{1}{c|}{ 83.5 } & \multicolumn{1}{c|}{ 72.5 } & \multicolumn{1}{c|}{ 77.5 } & \multicolumn{1}{c|}{ 60.3 } & \multicolumn{1}{c|}{ 39.3 } & \multicolumn{1}{c|}{ 36.3 } & \multicolumn{1}{c}{ 37.8 }         \\

\multicolumn{2}{l|}{DeOP~\cite{han2023zero}}  & \multicolumn{1}{l|}{One-Stage}     & \multicolumn{1}{c|}{ - } & \multicolumn{1}{c|}{ 88.2 } & \multicolumn{1}{c|}{ 74.6 } & \multicolumn{1}{c|}{ 80.8 } & \multicolumn{1}{c|}{ - } & \multicolumn{1}{c|}{ 38.0 } & \multicolumn{1}{c|}{ 38.4 } & \multicolumn{1}{c}{ 38.2 }         \\

\multicolumn{2}{l|}{ZegCLIP~\cite{zhou2022zegclip}}  & \multicolumn{1}{l|}{One-Stage}     & \multicolumn{1}{c|}{ 94.6 } & \multicolumn{1}{c|}{ 91.9 } & \multicolumn{1}{c|}{ 77.8 } & \multicolumn{1}{c|}{ 84.3 } & \multicolumn{1}{c|}{ 62.0 } & \multicolumn{1}{c|}{ 40.2 } & \multicolumn{1}{c|}{ 41.4 } & \multicolumn{1}{c}{ 40.8 } \\ \hline

\multicolumn{2}{l|}{LDVC (Ours)}   & \multicolumn{1}{l|}{One-Stage}     & \multicolumn{1}{c|}{ \textbf{95.5} } & \multicolumn{1}{c|}{ \textbf{92.5} } & \multicolumn{1}{c|}{ \textbf{82.3} } & \multicolumn{1}{c|}{ \textbf{87.2} } & \multicolumn{1}{c|}{ \textbf{64.3} } & \multicolumn{1}{c|}{ \textbf{43.2} } & \multicolumn{1}{c|}{ \textbf{45.0} } & \multicolumn{1}{c}{ \textbf{44.1} } \\ \bottomrule
\end{tabular}
}}
\caption{Comparison with SOTA methods under "inductive" GZS3 setting on VOC 2012 and COCO-Stuff 164k datasets.}
     \label{tab:compare-sota}
\vspace{-0.5em}
\end{table*}
\vspace{-1em}
\section{Experiment}

\begin{figure*}[!t]
\centering
        \includegraphics[width=0.98\textwidth]{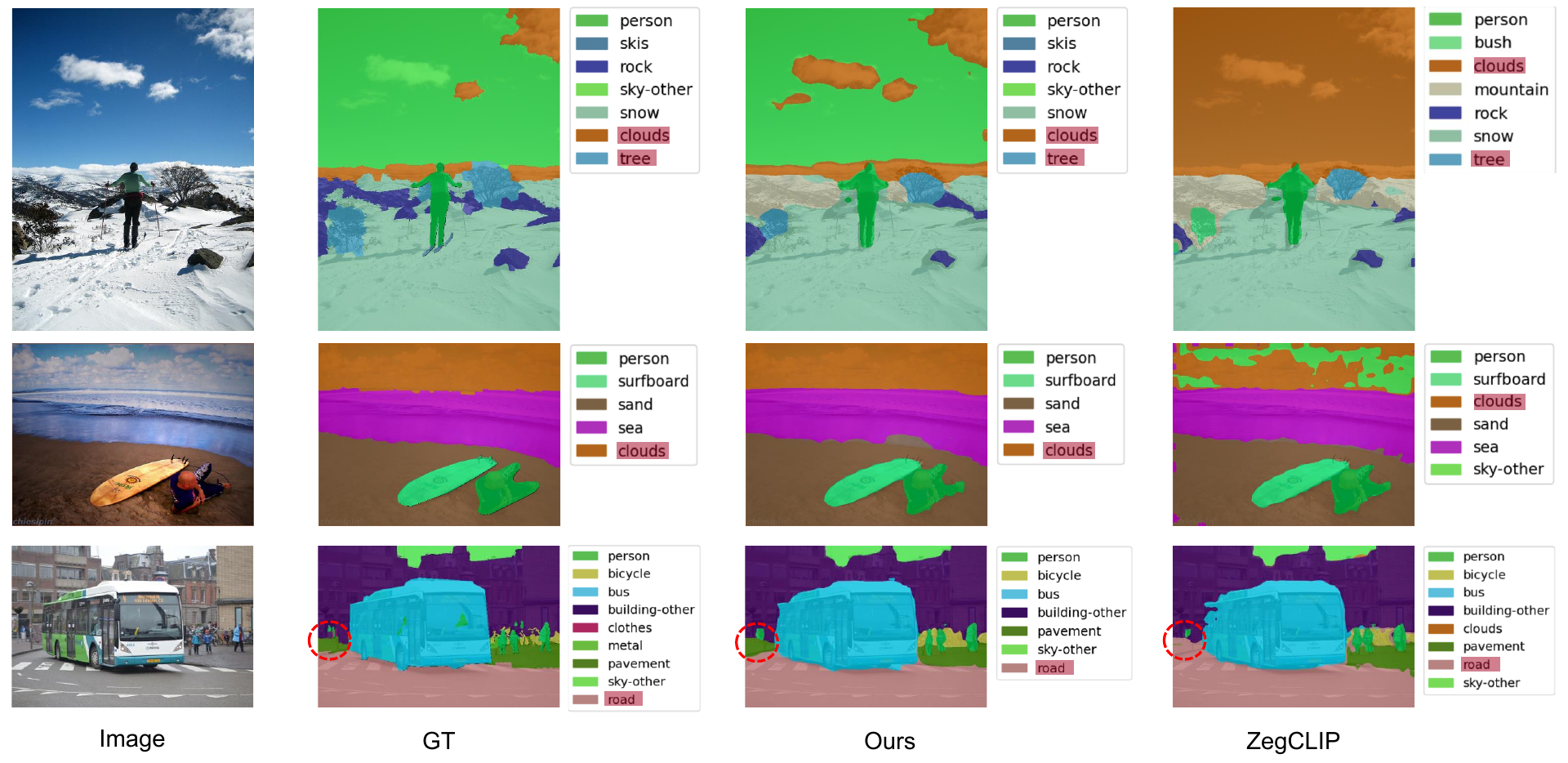}
\vspace{-0.3em}
\caption{\textbf{Visualization of segmentation prediction under inductive settings on COCO-Stuff 164k.} We compare the segmentation results between our approach and the SOTA method. The categories with color marked are unseen classes and others are seen classes.}
\vspace{-1em}
\label{fig:visual}
\end{figure*}

\subsection{Setting}
\label{par:setting}
\noindent\textbf{Datasets.} We evaluate our proposed method on two widely used datasets: VOC2012~\cite{everingham2011pascal} and COCO-Stuff 164K~\cite{caesar2018cvpr}. VOC2012 contains 10,582 augmented images for training and 1,449 for validation. Following previous works~\cite{zhou2022zegclip,ding2022decoupling}, we delete "background" class and use 15 classes as seen classes and 5 classes as unseen classes. COCO-Stuff 164K includes all 164K images from COCO~\cite{lin2014microsoft}, has 171 named categories and contains 118K images for training, 5K for evaluation. We use 156 classes as seen classes and 15 classes as unseen classes. The unseen classes of each dataset are shown in supplementary. 

\noindent\textbf{Metrics.} Following previous methods, we use three metrics to evaluate our results: mIoU(S), mIoU(U) and hIoU. mIoU(S) and mIoU(U) represent the mean Intersection over Union (mIoU) on seen classes and unseen classes, respectively. hIoU denotes the harmonic mean value of mIoU on seen and unseen classes: 
\[
hIoU = \frac{2\times mIoU(S)\times mIoU(U)}{mIoU(S)+mIoU(U)}.
\]

\noindent\textbf{Implementation Details.} We implement our method based on the MMSegmentation~\cite{mmseg2020} framework. We choose CLIP ViT-B as our default vision and language backbones. The number of visual deep prompt tokens is set as 100 and 40 in each layer of CLIP for COCO-stuff 164k and VOC2012. The number of language deep prompt tokens is 6 in each layer on both datasets.
The number of our decoder block is 2. 
For local consensus self-attention, window size $n$ is 4$\times$4 total number of windows is 64 and the number of selected windows $m$ is 16. Following ZegCLIP, we linearly combine focal loss~\cite{lin2017focal} and dice loss~\cite{sudre2017generalised} as our training loss. The loss weights for focal loss and dice loss are 100.0 and 1.0, respectively. We crop each image into 512$\times$512 in both the training and evaluation stage. During training, we adopt AdamW~\cite{loshchilov2017decoupled} as our optimizer and set the initial learning rate as 2e-4. We use the polynomial learning rate scheduler. The number of iterations for COCO-Stuff 164k and VOC2012 is set as 80k and 20k, respectively. All experiments are conducted on 4 RTX3090 GPUs with the batch size of 16.

\subsection{Comparison with State-of-the-Arts}

To demonstrate the effectiveness of our method, we compare with previous the-state-of-art methods in Tab. \ref{tab:compare-sota} and also provide visualization results on the COCO-Stuff 164k dataset in Fig. \ref{fig:visual}. Both of them show the great generalization ability on unseen classes of our method.

\noindent\textbf{Inductive Settings.} First, we evaluate our method under inductive settings where the unseen class names are not known during training stage. From Tab. \ref{tab:compare-sota}, our method surpasses previous approaches on both VOC 2012 and COCO-Stuff 164k datasets. It outperforms DeOP~\cite{han2023zero} by 7.7\% and 6.4\% in mIoU(U) and hIoU on VOC 2012, and by 6.6\% and 5.9\% on COCO-Stuff 164k, respectively. Furthermore, our method exceeds the leading one-stage method ZegCLIP~\cite{zhou2022zegclip} by 4.5\% and 2.9\% in mIoU(U) and hIoU on VOC 2012, and by 3.6\% and 3.3\% on COCO-Stuff 164k.

\noindent\textbf{Visualization Results.} From Fig.~\ref{fig:visual}, we visualize some segmentation results from our method and the SOTA method on COCO-Stuff 164k validation dataset which contains more objects in a scene. The categories marked with red are unseen classes. From Fig.~\ref{fig:visual}, we can see that comparing with ZegCLIP, our method recognizes objects in the images more accurately and achieves a higher match with the ground truth classes. In addition, our method is more accurate in predicting unseen class masks, for example, as the red circled part in the bottom row of Fig.~\ref{fig:visual} shown, our method successfully distinguishes between ``road'' and ``pavement'', while ZegCLIP regards ``pavement'' as ``road''. And on the top of Fig.~\ref{fig:visual}, it illustrates that ZegCLIP segments the whole upper region of the image as the ``clouds'' category, while our method produces more fine-grained results that successfully distinguish between ``clouds'' and ``sky-other'', which is even more accurate than the ground truth annotation. 

\noindent\textbf{Transductive Settings.} We also evaluate our method under transductive settings where the unseen class names are known during training stage. Thus, besides fine-tuning, this setting involves another self-tuning (ST) stage. We follows the same training strategy previous method~\cite{xu2022simple,zhou2022zegclip,kim2023zegot}. From Tab.~\ref{tab:transductive settings}, we compare our method with ZegOT and ZegCLIP on COCO-Stuff 164k datasets. Our method outperforms ZegOT by 2.9\% and 1.1\% on mIoU(S) and mIoU(U), respectively. Moreover our method exceeds ZegCLIP by 0.5\% and 0.4\% on mIoU(S) and mIoU(U), respectively. The results demonstrate that our method also performs well under transductive settings.

\noindent\textbf{Fully supervised.} We also provide the results of fully supervised semantic segmentation on COCO-Stuff 164k, which serves as the upper bound of our approach. In Tab.~\ref{tab:finetune-settings}, compared to ZegClip with 14M learnable parameters and ZegOT with 21M learnable parameters, our model has fewer parameters, specifically 11M. With less learnable parameters, our model still outperforms ZegOT and ZegCLIP by 5.3\% and 3.7\% on mIoU, respectively. These results demonstrate that our approach can better enhance the segmentation ability of CLIP’s backbone with less learnable parameters.

\begin{table}[!t]
\centering
\setlength{\tabcolsep}{0.7em}
    {\renewcommand{\arraystretch}{1.1}
\begin{tabular}{cc|cl|cccc|cccc}
\toprule

\multicolumn{1}{c|}{Method} &\multicolumn{1}{c|}{mIoU(S)} & \multicolumn{1}{c|}{\textbf{mIoU(U)}}  & \multicolumn{1}{c}{hIoU} \\ \hline

\multicolumn{1}{c|}{ZegOT~\cite{kim2023zegot}+ST} & \multicolumn{1}{c|}{38.2} & \multicolumn{1}{c|}{59.2}    & \multicolumn{1}{c}{46.4} \\

\multicolumn{1}{c|}{ZegCLIP~\cite{zhou2022zegclip}+ST} & \multicolumn{1}{c|}{40.6} & \multicolumn{1}{c|}{59.9}    & \multicolumn{1}{c}{48.4} \\

\multicolumn{1}{c|}{Ours+ST} &\multicolumn{1}{c|}{\textbf{41.1}} & \multicolumn{1}{c|}{\textbf{60.3}}    & \multicolumn{1}{c}{\textbf{48.8}} \\

\bottomrule
\end{tabular}
}
\vspace{-0.5em}
\caption{Comparison with SOTA methods under transductive ZS3 setting on COCO Stuff164k dataset. ``ST'' represents self-training.}
     \label{tab:transductive settings}
\vspace{-0.5em}
\end{table}

\begin{table}[!t]
\centering
\resizebox{0.99\linewidth}{!}{%
\setlength{\tabcolsep}{0.7em}
    {\renewcommand{\arraystretch}{1.1}
\begin{tabular}{cc|cl|cccc|cccc}
\toprule

\multicolumn{1}{c|}{Method} &\multicolumn{1}{c|}{Params
}&\multicolumn{1}{c|}{mIoU} &\multicolumn{1}{c|}{mIoU(S)} & \multicolumn{1}{c|}{\textbf{mIoU(U)}}  & \multicolumn{1}{c}{hIoU} \\ \hline

\multicolumn{1}{c|}{ZegOT~\cite{kim2023zegot}} & \multicolumn{1}{c|}{21M} &\multicolumn{1}{c|}{40.1} &\multicolumn{1}{c|}{38.3} & \multicolumn{1}{c|}{58.7}    & \multicolumn{1}{c}{46.4} \\

\multicolumn{1}{c|}{ZegCLIP~\cite{zhou2022zegclip}} & \multicolumn{1}{c|}{14M} &\multicolumn{1}{c|}{42.7}&\multicolumn{1}{c|}{40.7} & \multicolumn{1}{c|}{63.2}    & \multicolumn{1}{c}{49.6} \\

\multicolumn{1}{c|}{Ours} & \multicolumn{1}{c|}{11M} & \multicolumn{1}{c|}{\textbf{46.4}}&\multicolumn{1}{c|}{\textbf{44.5}} & \multicolumn{1}{c|}{\textbf{65.9}}    & \multicolumn{1}{c}{\textbf{53.1}} \\

\bottomrule
\end{tabular}
}}
\vspace{-0.5em}
\caption{Comparison with SOTA methods under fully supervised semantic segmentation on COCO-Stuff 164k dataset, which serves as the upper bound of ZS3 methods.}
     \label{tab:finetune-settings}
\vspace{-1.0em}
\end{table}
\subsection{Ablation Study}

All ablation experiments are conducted on COCO-Stuff 164k validation dataset.

\noindent\textbf{Effectiveness of VLPT and LCTD.} 
The effectiveness of the Vision-language Prompt Tuning (VLPT) and Local Consensus Transformer Decoder(LCTD) is introduced first for the ablation study, as shown in Table \ref{tab:VLPTLCTD}. The hIoU is improved from 40.8\% to 42.2\% when VLPT is used with the baseline ZegCLIP~\cite{zhou2022zegclip}. 

One main difference between the proposed LCTD decoder and SegViT decoder in ZegCLIP lies in that we choose image features as query and class embeddings as key and value in cross attention of the transformer decoder. As we illustrated in Sec.~\ref{introduce:explain}, under the parameter-free tuning and visual prompt tuning, the updating of visual space is limited, leading to a less structured visual space. This will increase the difficulty of alignment, thereby affecting the segmentation results. Equipped with LCTD without Local Consensus Self-Attention (LCSA), the mIoU(U) gets a gain of 2.3\%. It further increases to 45.0\% when LCSA is utilized.


\noindent\textbf{Ablation Studies on VLPT.} 
The ablation studies on VLPT are detailed including prompt-tuning strategies, initialization of the language prompt-tuning, and the lengths of language prompt tokens. 

We compare different prompt tuning methods first, as shown in Tab.~\ref{tab:prompt methods}. It illustrates that applying deep prompt tuning on both vision and language backbone simultaneously brings the best performance on both seen and unseen classes. If deep prompt tuning is only applied to the language backbone, the performance will decrease 2.28\% and 5.54\% on seen and unseen classes, respectively. If deep prompt tuning is only applied to the vision backbone, the performance will decrease 0.94\% and 1.92\%, respectively. The results also show that prompt tuning on the vision backbone has more effects on segmentation performance.

In Tab.~\ref{tab:prompt initialize}, we compare different prompt initialization methods on the language backbone. ``Random'' means we use random Gaussian distribution to initialize prompts and ``pretrain'' means we use the initial strategy mentioned in Sec~\ref{par:VLPT}. From the results, the proposed "pretrain" initialization is better than random initialization, especially prominent in unseen classes, where it achieves 4.5\% mIoU gain compared to the random initialization. 

\begin{table}[!t]
\centering
\setlength{\tabcolsep}{0.55em}
    {\renewcommand{\arraystretch}{1.1}
\begin{tabular}{ccc|c|c|c}
\toprule
VLPT & LCTD & LCSA & mIoU(S) & mIoU(U) & hIoU \\ 
\hline
 & Baseline  &   & 40.2 & 41.4 & 40.8 \\
\hline
\checkmark  &   &   & 42.3 & 41.9 & 42.2 \\
\checkmark &  \checkmark & & 42.8                    & 44.2 & 43.5 \\
\rowcolor{gray!20}
\checkmark &  \checkmark & \checkmark & \textbf{43.2} & \textbf{45.0} & \textbf{44.1} \\
\bottomrule
\end{tabular}
}
\vspace{-1.0em}
\caption{Effectiveness of VLPT, LCTD and LCSA.}
     \label{tab:VLPTLCTD}
\vspace{-0.5em}
\end{table}

\begin{table}[!t]
\centering
\setlength{\tabcolsep}{0.6em}
    {\renewcommand{\arraystretch}{1.1}
\begin{tabular}{cc|cl|cccc|cccc}
\toprule
\multicolumn{1}{c|}{Prompt Tuning Method} & \multicolumn{1}{c|}{mIoU(S)} & \multicolumn{1}{c|}{\textbf{mIoU(U)}}  & \multicolumn{1}{c}{hIoU} \\ \hline
\multicolumn{1}{c|}{Language-Only} & \multicolumn{1}{c|}{40.92} & \multicolumn{1}{c|}{39.48}    & \multicolumn{1}{c}{40.19} \\
\multicolumn{1}{c|}{Vision-Only} & \multicolumn{1}{c|}{42.26} & \multicolumn{1}{c|}{43.10}    & \multicolumn{1}{c}{42.68} \\
\rowcolor{gray!20}
\multicolumn{1}{c|}{Vision-Language} & \multicolumn{1}{c|}{\textbf{43.20}} & \multicolumn{1}{c|}{\textbf{45.02}}    & \multicolumn{1}{c}{\textbf{44.09}} \\
\bottomrule
\end{tabular}
}
\vspace{-1em}
\caption{Ablation of different prompt tuning on CLIP's encoders.}
     \label{tab:prompt methods}
\vspace{-0.5em}
\end{table}

\begin{table}[!t]
\centering
\setlength{\tabcolsep}{0.7em}
    {\renewcommand{\arraystretch}{1.1}
\begin{tabular}{cc|cl|cccc|cccc}
\toprule
\multicolumn{1}{c|}{Prompt Initialization}  & \multicolumn{1}{c|}{mIoU(S)} & \multicolumn{1}{c|}{\textbf{mIoU(U)}}  & \multicolumn{1}{c}{hIoU} \\ \hline
\multicolumn{1}{c|}{Random}  & \multicolumn{1}{c|}{43.1} & \multicolumn{1}{c|}{40.5}  & \multicolumn{1}{c}{41.8} \\
\rowcolor{gray!20}
\multicolumn{1}{c|}{Pretrain}  & \multicolumn{1}{c|}{\textbf{43.2}} & \multicolumn{1}{c|}{\textbf{45.0}}  & \multicolumn{1}{c}{\textbf{44.1}} \\
\bottomrule
\end{tabular}
}
\vspace{-1em}
\caption{Ablation of different language prompt initialization strategies.}
\label{tab:prompt initialize}
\vspace{-1em}
\end{table}



\begin{table*}[!t]
\centering
\resizebox{0.99\linewidth}{!}{%
\setlength{\tabcolsep}{0.7em}
    {\renewcommand{\arraystretch}{1.1}
\begin{tabular}{cc|cl|cccc|cccc}
\toprule
\multicolumn{1}{l|}{\multirow{2}{*}{Different ZS3 decoders}} & \multicolumn{3}{c|}{Frozen} &\multicolumn{3}{c|}{Visual Prompt-tuning} & \multicolumn{3}{c}{Vision-Language Prompt Tuning}
\\ \cline{2-10} 

\multicolumn{1}{l|}{}    & \multicolumn{1}{c|}{mIoU(S)}          & \multicolumn{1}{c|}{\textbf{mIoU(U)}} &\multicolumn{1}{c|}{hIoU}  & \multicolumn{1}{c|}{mIoU(S)}          & \multicolumn{1}{c|}{\textbf{mIoU(U)}} &\multicolumn{1}{c|}{hIoU}
& \multicolumn{1}{c|}{mIoU(S)}          & \multicolumn{1}{c|}{\textbf{mIoU(U)}} &\multicolumn{1}{c}{hIoU}
\\ \hline

\multicolumn{1}{l|}{ZegCLIP~\cite{zhou2022zegclip}}    & \multicolumn{1}{c|}{32.3}          & \multicolumn{1}{c|}{32.5} &\multicolumn{1}{c|}{32.4}  & \multicolumn{1}{c|}{40.2} & \multicolumn{1}{c|}{41.4}  & \multicolumn{1}{c|}{40.8}
& \multicolumn{1}{c|}{42.3}          & \multicolumn{1}{c|}{41.9} &\multicolumn{1}{c}{42.2} \\ 
\multicolumn{1}{l|}{LDVC w/o LCSA}    & \multicolumn{1}{c|}{38.7} & \multicolumn{1}{c|}{39.5}  & \multicolumn{1}{c|}{39.1}  & \multicolumn{1}{c|}{41.7} & \multicolumn{1}{c|}{42.8}  & \multicolumn{1}{c|}{42.2}
& \multicolumn{1}{c|}{42.8} & \multicolumn{1}{c|}{44.2}  & \multicolumn{1}{c}{43.5} \\ 
\rowcolor{gray!20}
\multicolumn{1}{l|}{LDVC}    & \multicolumn{1}{c|}{39.8} & \multicolumn{1}{c|}{41.1} & \multicolumn{1}{c|}{40.4}  & \multicolumn{1}{c|}{42.3} & \multicolumn{1}{c|}{43.1}    & \multicolumn{1}{c|}{42.7}
& \multicolumn{1}{c|}{\textbf{43.2}} & \multicolumn{1}{c|}{\textbf{45.0}}  & \multicolumn{1}{c}{\textbf{44.1}} \\ 
\bottomrule
\end{tabular}
}}
\caption{Comparison with different ZS3 decoders under parameter-free fine-tuning and different prompt tuning settings.}
     \label{tab:different decoders}
\vspace{-1em}
\end{table*}

We also verify the VLPT used as a plug-in to the baseline, as the first row shown in Table~\ref{tab:different decoders}.
In Table~\ref{tab:different decoders}, it also verifies the LCTD by column comparison. The performances of mIoU(S), mIoU(U), and hIoU are all increased when the CLIP is frozen, tuned with the visual prompt, or the visual-language prompt.

\noindent\textbf{Ablation Studies on LCTD.} 
Next, we give the ablation studies on LCTD about the number of selected windows in local consensus self-attention and the number of stacked decoder blocks.

\begin{table}[!t]
\centering
\setlength{\tabcolsep}{0.7em}
    {\renewcommand{\arraystretch}{1.1}
\begin{tabular}{cc|cl|cccc|cccc}
\toprule

\multicolumn{1}{c|}{TopK} & \multicolumn{1}{c|}{mIoU(S)} & \multicolumn{1}{c|}{\textbf{mIoU(U)}}  & \multicolumn{1}{c}{hIoU} \\ \hline
\multicolumn{1}{c|}{8} & \multicolumn{1}{c|}{43.14} & \multicolumn{1}{c|}{44.87}    & \multicolumn{1}{c}{43.99} \\
\rowcolor{gray!20}
\multicolumn{1}{c|}{16} & \multicolumn{1}{c|}{\textbf{43.20}} & \multicolumn{1}{c|}{\textbf{45.02}}    & \multicolumn{1}{c}{\textbf{44.09}} \\
\multicolumn{1}{c|}{32} & \multicolumn{1}{c|}{43.20} & \multicolumn{1}{c|}{45.00}    & \multicolumn{1}{c}{44.08} \\

\multicolumn{1}{c|}{48} & \multicolumn{1}{c|}{42.98} & \multicolumn{1}{c|}{44.34}    & \multicolumn{1}{c}{43.64} \\

\bottomrule
\end{tabular}
}
\vspace{-0.8em}
\caption{Ablation of the number of selected windows in local consensus self-attention.}
\label{tab:number of selected windows}
\vspace{-0.8em}
\end{table}


\begin{table}[!t]
\centering
\setlength{\tabcolsep}{0.7em}
    {\renewcommand{\arraystretch}{1.1}
\begin{tabular}{cc|cl|cccc|cccc}
\toprule

\multicolumn{1}{c|}{Number of Blocks} & \multicolumn{1}{c|}{mIoU(S)} & \multicolumn{1}{c|}{\textbf{mIoU(U)}}  & \multicolumn{1}{c}{hIoU} \\ \hline
\rowcolor{gray!20}
\multicolumn{1}{c|}{2} & \multicolumn{1}{c|}{43.20} & \multicolumn{1}{c|}{\textbf{45.02}}    & \multicolumn{1}{c}{\textbf{44.09}} \\
\multicolumn{1}{c|}{3} & \multicolumn{1}{c|}{43.06} & \multicolumn{1}{c|}{44.67}    & \multicolumn{1}{c}{43.85} \\
\multicolumn{1}{c|}{4} & \multicolumn{1}{c|}{\textbf{44.01}} & \multicolumn{1}{c|}{43.26}    & \multicolumn{1}{c}{43.63} \\

\bottomrule
\end{tabular}
}
 \vspace{-0.8em}
\caption{Ablation of the number of decoder
blocks.}
\label{tab:number of decoder blocks}
\vspace{-1.2em}
\end{table}



In Tab.~\ref{tab:number of decoder blocks}, we evaluate the impact of the number of decoder blocks on ZS3 performance. A lightweight local consensus transformer decoder with two block is already sufficient.
The performances are similar between a two-layer decoder and a three-layer decoder. Increasing the number of layers leads to a decrease in mIoU(U). Considering both the number of parameters and performance, we stack two decoder blocks. This choice also results in our approach having fewer learnable parameters compared to ZegCLIP, with a total of 11M learnable parameters.

\begin{figure}[!t]
\centering
        \includegraphics[width=0.99\linewidth]{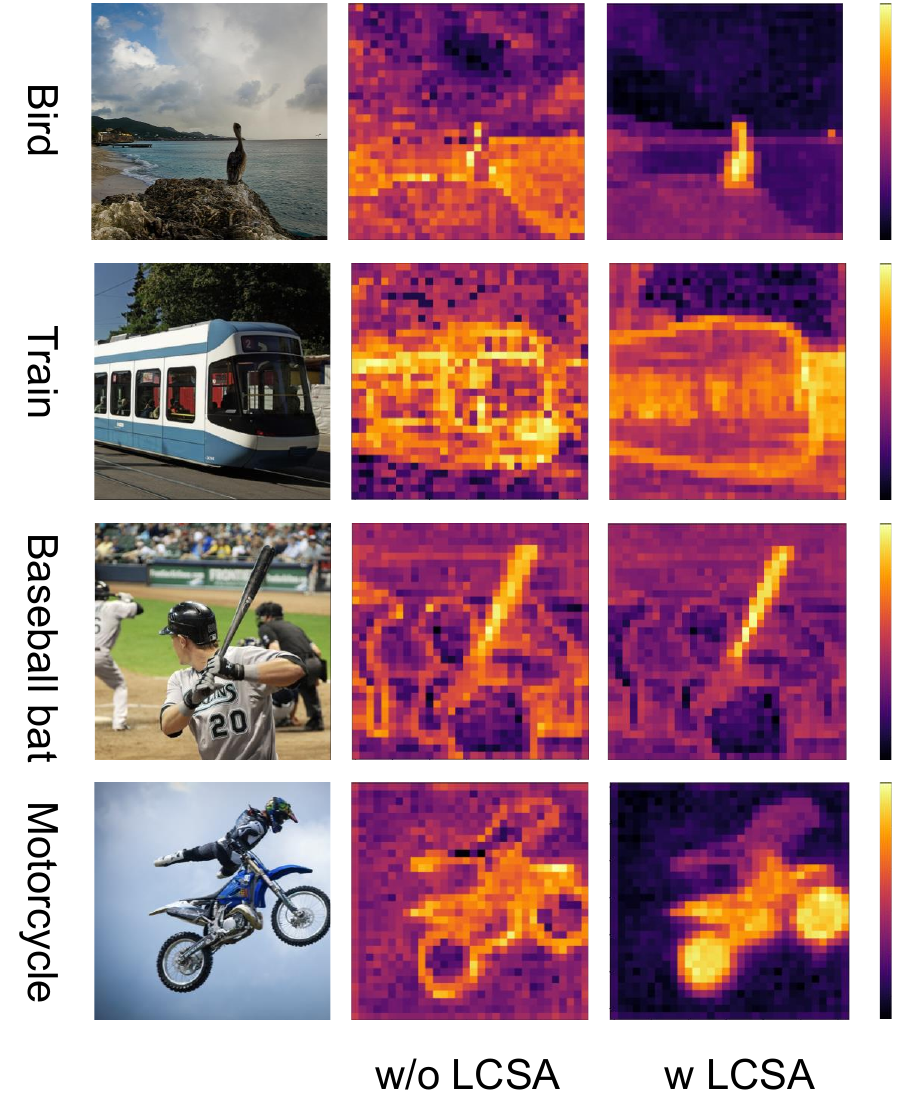}
\vspace{-1em}
\caption{Visualization of cross attention between image features and class embeddings in our decoder with LCSA or without LCSA.}
\vspace{-1em}
\label{fig:intermediate cross attention map}
\vspace{-1em}
\end{figure}

In our local consensus transformer decoder, we replace vanilla self-attention with local consensus self-attention to enhance the local semantic consensus in the same object. 
In Fig.~\ref{fig:intermediate cross attention map}, we visualize the cross attention map in the last block of our decoder. We observe that after incorporating local consensus self-attention, the noise in the cross-attention map is reduced, and the semantics within the same object become more consistent, which demonstrate the effectiveness of local consensus self-attention. Besides, we give the ablation study about the number of selected windows in local consensus self-attention in Tab.~\ref{tab:number of selected windows}. The total number of windows in patch features is 64. In Tab.~\ref{tab:number of selected windows}, the results show the constant performance when the number is 16 and 32. However, when the number is 48, the local consensus self-attention starts behaving like vanilla self-attention, gathering too many irrelevant patches, leading to a performance decline.

\section{Conclusion}

We propose a language-driven visual consensus approach for zero-shot semantic segmentation. To solve the issues about overfitting on seen classes and samll fragmentation in segmentation masks, we propose a new local consensus transformer decoder which adopts image features as query and class embeddings as key and value in cross attention. To further enhance the semantic consistency of image features in the same object, we introduce route attention mechanism to vanilla self-attention of our decoder. Besides, under a visual-language prompting strategy, our approach further improves segmentation capabilities of CLIP. Extensive experiments have demonstrated the effectiveness of our method. Hope our work can inspire future research.

{
    \small
    \bibliographystyle{ieeenat_fullname}
    \bibliography{main}
}

\end{document}